\documentclass{article}

\usepackage{arxiv}

\usepackage[utf8]{inputenc} % allow utf-8 input
\usepackage[T1]{fontenc}    % use 8-bit T1 fonts
\usepackage{hyperref}       % hyperlinks
\usepackage{url}            % simple URL typesetting
\usepackage{booktabs}       % professional-quality tables
\usepackage{amsfonts}       % blackboard math symbols
\usepackage{nicefrac}       % compact symbols for 1/2, etc.
\usepackage{microtype}      % microtypography
\usepackage{lipsum}
\usepackage{graphicx}

\usepackage{framed,multirow}

\usepackage{subfig}

\usepackage{float}

\title{Personality Perception in Human Videos Altered by Motion Transfer Networks}

\newcommand*\samethanks[1][\value{footnote}]{\footnotemark[#1]}

\author{
  Ayda Yurtoğlu \thanks{Joint first authors.} \\
  Department of Computer Engineering \\
  Bilkent University, Ankara, Turkey \\
  \texttt{ayda.yurtoglu@ug.bilkent.edu.tr} \\
  \And
  Sinan Sonlu \samethanks \\
  Department of Computer Engineering \\
  Bilkent University, Ankara, Turkey \\
  \texttt{sinan.sonlu@bilkent.edu.tr} \\
  \And
  Yalım Doğan \samethanks \\
  Department of Computer Engineering \\
  Bilkent University, Ankara, Turkey \\
  \texttt{yalim.dogan@bilkent.edu.tr} \\
  \And
  Uğur Güdükbay \thanks{Corresponding author: Tel.: +90-312-290-1386; fax: +90-312-266-4047} \\
  Department of Computer Engineering \\
  Bilkent University, Ankara, Turkey \\
  \texttt{gudukbay@cs.bilkent.edu.tr} \\
}

\begin{document}
\maketitle
\begin{abstract}
The successful portrayal of personality in digital characters improves communication and immersion. Current research focuses on expressing personality through modifying animations using heuristic rules or data-driven models. While studies suggest motion style highly influences the apparent personality, the role of appearance can be similarly essential. This work analyzes the influence of movement and appearance on the perceived personality of short videos altered by motion transfer networks. We label the personalities in conference video clips with a user study to determine the samples that best represent the Five-Factor model's high, neutral, and low traits. We alter these videos using the Thin-Plate Spline Motion Model, utilizing the selected samples as the source and driving inputs. We follow five different cases to study the influence of motion and appearance on personality perception. Our comparative study reveals that motion and appearance influence different factors: motion strongly affects perceived extraversion, and appearance helps convey agreeableness and neuroticism.
\end{abstract}

% keywords can be removed

\section{Introduction}
Personality investigates the qualities that form the individual. The Five-Factor model~\cite{mccrae2005personality} examines personality under five orthogonal dimensions: openness to experience, conscientiousness, extraversion, agreeableness, and neuroticism. 
An accurate portrayal of these factors helps improve immersion and communication~\cite{durupinar2016perform, smith2017understanding, sonlu2021conversational}; applications utilize motion~\cite{vinciarelli2014survey} and physical appearance~\cite{naumann2009personality} features to convey different personalities. While both approaches are well studied in virtual characters separately, the interaction between appearance and movement remains to be explored, especially in videos altered by motion transfer networks that enable a practical way for character animation. This work analyzes the impact of appearance and motion in conveying different personalities of the Five-Factor model in videos altered by motion transfer networks.

Expressive systems utilize a high-level understanding of human motion to define heuristics-based rules that control the virtual human's animation and personality~\cite{durupinar2016perform}. Such high-level features also apply to motion~\cite{KimEtAl22} and personality~\cite{erkocc2022skeleton} analysis. Appearance-based features, on the other hand, vary dramatically: Hair color~\cite{lawson1971hair}, hair length~\cite{pancer1978length}, facial shape, skin texture, and viewing angle~\cite{jones2012signals}, body shape~\cite{hu2018first, namatame2016personality}, clothing~\cite{feinberg1992clothing}, and eye size~\cite{paunonen1999facial} can influence apparent personality. Applications generally focus on expressing personality in predetermined characters and thus do not aim to alter the appearance. In an animated film, different scenes may require the same character to express different personality traits without a change in physical appearance; however, the general appearance is highly influential on perceived personality~\cite{naumann2009personality}.

Suppose we have two characters of different appearances. Would adopting the same motion expressing extraversion similarly influence their apparent personality? Which would be more influential if the appearance expresses extraversion and the movements express the opposite? These questions inspired us to investigate the interaction of appearance- and motion-based cues in videos conveying personality.

With the rise of deep generative models~\cite{sha2021deep, kammoun2022generative, tewari2022advances}, the future of CGI can be data-driven. While the ownership of artificial creativity is still in debate~\cite{eshraghian2020human}, studies utilize data-driven methods to synthesize the motion~\cite{ye2022human, he2022evaluating} and the looks of the virtual characters~\cite{toshpulatov2021generative}. Most of our digital characters could adopt such generative methods in the future, and well-representing different personalities in virtual human characters would be essential for a more diverse digital world; we analyze the capabilities of a motion transfer network to modify human videos to express different personalities as a step towards this goal.

In a user study, we label the apparent personalities of the individuals in the TED-talks dataset~\cite{siarohin2021motion} using Ten-Item Personality Inventory~\cite{gosling2003very}. We pick the samples representing high, neutral, and low traits for the five personality dimensions. We use the Thin-Plate Spline Motion Model~(TPS)~\cite{zhao2022thinplate} to generate new sequences with the selected samples, using them as the source and driving inputs. The output videos have the source's appearance and the driving sample's motion. We examine five different cases to examine the influence of different input combinations.

We list our research questions as follows:

\begin{itemize}
    \item Can motion transfer networks alter personality perception in human videos?
    \item Do appearance and motion-based cues similarly impact personality expression?
    \item Do adjustments made by motion transfer networks influence each personality factor similarly?
\end{itemize}

This work's contribution is an extensive analysis of TPS's performance in portraying personality through utilizing different samples as the source and driving inputs. Our framework, data, and user study results are available on our GitHub repository~\footnote{\url{https://github.com/sinansonlu/motion-transfer-personality}.} for further studies. The results suggest that using the corresponding high-trait samples as driving input helps portray extraversion and openness; using them as the source input helps express agreeableness and neuroticism. The influence of motion transfer on perceived conscientiousness is limited.

\section{Related Work}

Theories emphasize different elements as the source of personality~\cite{boyle2008personality}; biological approaches focus on the link between genetics and apparent traits~\cite{davidson2001toward}, behavioral approaches explain the causes through interaction with the environment~\cite{kathleen2020behavioral}, and psychodynamic approaches highlight the unconscious mind~\cite{bornstein2012psychodynamic}. Although personality models investigate similar human traits, their categorizations vary~\cite{zuckerman1993comparison}. Among the many models used in psychology, the Five-Factor model~\cite{mccrae2005personality} is popular in computer science due to its orthogonal dimensions.

% The Five-Factor model examines the individual's personality under five orthogonal dimensions that form the acronym OCEAN.

% Openness to experience represents the intellectual aspect of personality, including traits such as insightfulness and curiosity. Conscientiousness describes thoughtfulness and goal-directed behavior; highly conscientious people are organized and mindful of details. Extraversion reflects seeking social connections and interaction; people high in extraversion are outgoing and comfortable in social situations. Agreeableness measures the degree of compassion and cooperation; highly agreeable people feel more empathy and concern for others. Neuroticism explains the tendency towards moodiness and emotional instability; highly neurotic people feel more stress and anxiety.

Both static cues, such as hair and clothing style, and dynamic cues, like facial expression and posture, signal personality-relevant information~\cite{agnew1984effect, naumann2009personality}. People's judgment of others' personalities depends on such cues in first encounters. Research indicates peer and self-reported personalities mostly agree for extraversion~\cite{funder1988friends} and vary in accuracy for other factors~\cite{ready2000self, pozzebon2009personality, kandler2010sources}. Accurate judgment depends on both parties; the peer should observe the cues correctly, and the self should express them clearly~\cite{funder2012accurate}. In computer-generated imagery, faithful representation of the cues is essential for expressing the desired personality. The successful portrayal of different personality traits in virtual humans can build trust~\cite{zhou2019trusting}, improve social presence~\cite{sajjadi2019personality} and evoke empathy~\cite{zibrek2018effect}. Consequently, many applications that involve human characters desire an appropriate portrayal of the personality.

\subsection{Expressing Personality}
Studies utilize expressive motion to create more human-like virtual characters. Appropriate gesturing improves realism and engagement~\cite{ferstl2021expressgesture}. Data-driven models can generate hand gestures accompanying input speech~\cite{ghorbani2022exemplar}. Appropriate use of facial expressions~\cite{milcent2022using} and nonverbal communication cues~\cite{thakkar2022understanding} induce empathy in users. Facial expressions reveal the individual's feelings with high accuracy~\cite{gloor2022your}, and the magnitude of facial motion in animated characters influences personality perception~\cite{HydeEtAl16}; consequently, personality recognition systems often input facial~\cite{suen2019tensorflow} and body related~\cite{hu2023first} features together. Movement qualities such as smoothness, energy, and impulsivity correlate with apparent personality~\cite{GiraudEtAl15}. Body shape and motion style influence the perceived sex of virtual characters~\cite{McDonnellEtAl09}, causing users to behave differently to them~\cite{ZibrekEtAl20}.

Even with minimal information, motion indicators accurately predict personality~\cite{koppensteiner2013motion}, inspiring studies to utilize gesturing~\cite{neff2010evaluating, smith2017understanding} and hand motion~\cite{WangEtAl16} in expressive animation. High-level motion features control the apparent personality~\cite{durupinar2016perform} and are useful in motion classification and emotion analysis~\cite{hachimura2005analysis, aristidou2015emotion, dewan2018spatio, ajili2019human}. Utilizing personality-specific dialogues, vocal features, and facial expressions improves the distinction of opposite personalities in conversational agents~\cite{sonlu2021conversational}; speech and animation realism influence the perceived personality~\cite{thomas2022investigating}.

Studies alter 3D face models~\cite{lang20193d}, use automatic gaze animation~\cite{durupinar2020personality} and synthetic body poses~\cite{calsius2019synthesizing} to express different personalities. Similar techniques apply to interactive humanoid robots~\cite{zabala2021expressing}. Imperfections in virtual characters can cause an uncanny effect~\cite{perez2020exploring} when they are very close to real but have noticeable divergence~\cite{mori2012uncanny}; this results in decreased trustworthiness~\cite{yuan2019crossing} and empathy~\cite{higgins2023investigating} towards expressive virtual characters. Positive traits like high agreeableness and emotional stability can weaken this negative effect~\cite{paetzel2021influence}. A recent study reveals that virtual characters' appearance and animation realism influence the perceived social presence and affinity while portraying different emotional facial expressions~\cite{nabila2023}.  Unlike existing research, this work examines the personality expression in human videos altered by motion transfer networks.

\subsection{Generating and Altering Human Videos}
There is an increasing interest in generating and altering human videos using data-driven models. Supervised methods in this field utilize 3D models~\cite{deng2020disentangled, geng20193d, thies2020face2face, doukas2021headgan, liu2019neural}, domain labels~\cite{choi2018stargan}, semantic segmentations~\cite{nirkin2019fsgan}, and landmarks~\cite{ha2019marionette, zakharov2019fewshot, chan2019everybody, ren2020deep, zhu2019progressive}. Generalizing these methods to different object categories is challenging and often requires fine-tuning~\cite{zakharov2019fewshot} or retraining~\cite{chan2019everybody}. Unsupervised image animation techniques address the limitations of the supervised methods and allow arbitrary subjects to be animated without prior knowledge, using a source and a driving input. For instance, Monkey-Net~\cite{siarohin2019animating} acquires motion information via self-learned key points and applies the corresponding trajectories to animate the source image. First Order Motion Model~(FOMM)~\cite{NEURIPS2019_31c0b36a} transfers complex motions to arbitrary objects by combining self-learned key points with local affine transformations, applying first-order Taylor expansions to approximate the motion near each key point; however, as local affine transformations are linear, they create limitations for objects that do not move linearly.

Motion Representations for Articulated Animation (MRAA) provides a warp-based method to identify object parts and extract their motion using PCA-based estimation~\cite{siarohin2021motion}. MRAA predicts the background motion separately to overcome the limitations of FOMM. These warp-based methods use optical flow estimators that warp the source image features to the driving video. Recycle-GAN~\cite{bansal2018recyclegan} utilizes a conditional GAN focusing on the video-to-video translation of arbitrary objects; unlike FOMM and MRAA, Recycle-GAN uses warp-free synthesis. MoCoGAN~\cite{tulyakov2017mocogan} is an unconditional GAN that controls the content and movement of the input videos. To address non-linear movements of the human body, TransMoMo~\cite{yang2020transmomo} is designed explicitly for representing articulated motion and disentanglements. Unlike the other data-driven models, Transformation-Synthesis Network~(TS-Net)~\cite{ni2022crossidentity} has a dual branch structure: warp-based transformation branch and warp-free synthesis branch, which are used together to generate video frames.

We use TPS~\cite{zhao2022thinplate}, a warp-based motion transfer framework that addresses the shortcomings of previous unsupervised methods. It can generate decent-quality videos even when a large pose gap exists between the source image and the driving video. It also overcomes the limitations of local affine transformations using TPS motion estimation. It demonstrates a better optical flow performance with a substantial inpainting capability than its ancestors.

\section{The Method and Experiments}
We first choose short video clips varying in motion and appearance from a dataset of conference talks. We label the personalities they represent in our first user study. Then, we identify the samples that best represent high, neutral, and low traits for each personality factor; we use them as TPS's source and driving inputs to generate new videos. Our second user study compares TPS's outputs following five cases to examine the influence of motion and appearance on personality perception.

\subsection{Choosing the Samples for Personality Questionnaire}
We manually label the available samples from the TED-Talks dataset~\cite{siarohin2021motion} based on the presenter's age, gender, race, clothing style (plain or complex), and motion (active or passive). We randomly picked 50 samples of unique label combinations to include in our first user study to better represent different groups and personalities. Among the 50 samples, 29 included a female speaker, 9 represented old age, 13 wore glasses, 17 had long hair, 2 had ties, 10 had facial hair, and 17 wore bright colors. Complete labels of the selected samples are available in our repository. Each sample is a muted sequence of 5-10 seconds from TED Talks. The videos focus on the presenter and only show the upper part of the body.

\subsection{User Studies}
We run two online user studies~\footnote{Bilkent University Ethical Committee for Human Research approves the studies with the decision number 2023\_06\_09\_01.} using our website implemented for this project. Participants could use any device supporting HTML and JavaScript to contribute to the user study. We acquire the participants among university students and their family members through email invitations; participation is entirely voluntary. Participants log in to the website using their email addresses and can resume the experiment anytime.

We did not prevent the users from participating in the second study if they participated in the first one, which enabled us to study both groups individually to analyze the influence of early exposure to the appearance and motion of the samples. Of all the 47 participants of the first study, 40 provided optional demographics (23 male and 17 female) with an average age of $34.45\pm12.93$. 11 of these users also participated in the second study. The total participant count for the second study is 30; 28 participants provided demographics (21 male and seven female) with an average age of $28.42\pm11.39$.

\subsection{First User Study}
In the first study, we show the selected 50 videos in random order one by one. Participants rate the individual's personality using sliders on a 7-point Likert scale. Each slider contains descriptive traits for opposing polarities of the personality factors from the Ten-Item Personality Inventory~(TIPI)~\cite{gosling2003very}, shown in Table~\ref{tab:opposite-traits}. We adopt this brief usage of TIPI from PERFORM's comparison study~\cite{durupinar2016perform}. We map the slider values to integers in the [-3, 3] range, where negative and positive values correspond to low and high traits of the corresponding personality factors, respectively. A value close to zero represents neutral traits.

\begin{table*}[htbp]
\small
\setlength{\tabcolsep}{3pt}
\caption{Summary of the Five-Factor personality dimensions and the corresponding high and low polarity keywords used in the first experiment.}
\centering
\begin{tabular}{llll}
\multicolumn{1}{c}{\textbf{Factor}} & \multicolumn{1}{c}{\textbf{Related Traits}} & \multicolumn{1}{c}{\textbf{High Polarity}} & \multicolumn{1}{c}{\textbf{Low Polarity}} \\ \hline
\textbf{Openness} & \begin{tabular}[c]{@{}l@{}}The intellectual aspect of the personality, includes traits\\ such as insightfulness, curiosity and self-examination.\end{tabular} & \begin{tabular}[c]{@{}l@{}}Open to new experiences\\ Complex\end{tabular} & \begin{tabular}[c]{@{}l@{}}Conventional\\ Uncreative\end{tabular} \\ \hline
\textbf{Conscientiousness} & \begin{tabular}[c]{@{}l@{}}Describes thoughtfulness and goal-directed behavior,\\ conscientious people are organized and mindful of details.\end{tabular} & \begin{tabular}[c]{@{}l@{}}Dependable\\ Self-disciplined\end{tabular} & \begin{tabular}[c]{@{}l@{}}Disorganized\\ Careless\end{tabular} \\ \hline
\textbf{Extraversion} & \begin{tabular}[c]{@{}l@{}}Reflects seeking social connections and interaction,\\ extraverts are outgoing and comfortable in social situations.\end{tabular} & \begin{tabular}[c]{@{}l@{}}Extraverted\\ Enthusiastic\end{tabular} & \begin{tabular}[c]{@{}l@{}}Reserved\\ Quiet\end{tabular} \\ \hline
\textbf{Agreeableness} & \begin{tabular}[c]{@{}l@{}}Measures the degree of compassion and cooperation,\\ agreeable people feel more empathy and concern for others.\end{tabular} & \begin{tabular}[c]{@{}l@{}}Sympathetic\\ Warm\end{tabular} & \begin{tabular}[c]{@{}l@{}}Critical\\ Quarrelsome\end{tabular} \\ \hline
\textbf{Neuroticism} & \begin{tabular}[c]{@{}l@{}}Explains the tendency towards moodiness and emotional\\ instability, neurotic people feel more stress and anxiety.\end{tabular} & \begin{tabular}[c]{@{}l@{}}Anxious\\ Easily upset\end{tabular} & \begin{tabular}[c]{@{}l@{}}Calm\\ Emotionally stable\end{tabular} \\ \hline
\end{tabular}
\label{tab:opposite-traits}
\end{table*}

We use one-way Analysis of Variance~(ANOVA) to investigate if the personality means of the samples differ significantly. Each personality factor is the dependent variable, and samples are the independent variables. Table~\ref{tab:anova-study-1} shows samples were rated significantly different for all personality factors. The extraversion variance is the highest, which can be due to this factor being the easiest to perceive. On the other hand, conscientiousness variance is the lowest; this is likely as all the samples are presenters, and thus they score relatively high in conscientiousness. 

\begin{table}[htbp]
\centering
\caption{ANOVA statistics per personality factor where samples are the independent variable and the participants' ratings are the dependent variable. Statistics marked with ** indicate $p < 0.001$.}
\begin{tabular}{clcccc}
\hline
\textbf{} & \multicolumn{1}{c}{\textbf{Source}} & \multicolumn{1}{c}{\textbf{SS}} & \multicolumn{1}{c}{\textbf{df}} & \multicolumn{1}{c}{\textbf{F}} & \multicolumn{1}{c}{\textbf{$\eta^{2}$}} \\ \hline
\multirow{2}{*}{O} & Between groups & 486.34 & 49 & 5.529** & 0.172 \\
 & Within groups & 2336.89 & 1302 & -- & -- \\ \hline
\multirow{2}{*}{C} & Between groups & 263.93 & 49 & 3.517** & 0.116 \\
 & Within groups & 1994.02 & 1302 & -- & -- \\ \hline
\multirow{2}{*}{E} & Between groups & 1222.64 & 49 & 12.753** & 0.324 \\
 & Within groups & 2547.32 & 1302 & -- & -- \\ \hline
\multirow{2}{*}{A} & Between groups & 866.91 & 49 & 8.943** & 0.251 \\
 & Within groups & 2575.54 & 1302 & -- & -- \\ \hline
\multirow{2}{*}{N} & Between groups & 609.13 & 49 & 5.589** & 0.173 \\
 & Within groups & 2895.49 & 1302 & -- & -- \\ \hline
\end{tabular}
\label{tab:anova-study-1}
\end{table}

Among our 50 samples, we choose $X_H$ and $X_L$ as the factors representing each personality factor's high and low polarity, respectively, where $X \in {O, C, E, A, N}$. We compare each sample pair based on their mean difference in each factor's score using Tukey Honestly Significant Differences~(HSD)~\cite{tukey1949comparing}; for each factor, the pair with the highest mean difference is selected as $X_H$ and $X_L$. For example, sample 45 and 20 have the highest mean extraversion difference. The mean extraversion of sample 45 is $2.28$, so it is selected as $E_H$, and sample 20 has a mean extraversion of $-1.75$, and consequently, it is selected as $E_L$. For each factor, we also choose a sample to represent the neutral traits $X_N$, where $X \in {O, C, E, A, N}$ such that $|{\textit{Mean}}(X_H) - {\textit{Mean}}(X_N)| + |{\textit{Mean}}(X_N) - {\textit{Mean}}(X_L)|$ is maximized, where the mean function calculates the mean score for the corresponding factor. We allow a sample to represent multiple factors; for example, sample 9 represents $O_L$ (low openness) and $N_N$ (neutral neuroticism). The index numbers of the samples correspond to their order in our dataset, so we replace them with $X_H$, $X_N$, and $X_L$ for the selected samples.

Table~\ref{tab:tukey-pairwise-study-1} shows the pairwise comparisons between the selected samples per factor. Although we are only interested in the matching personality measurement per sample pair, we report all factors' ratings. For example, $E_H$, $E_N$, and $E_L$ are selected to represent extraversion, but we also report their differences for the remaining factors. The results show that the selected $X_H-X_L$ pairs are most suitable to represent their matching factors but can also have significant differences for other factors. For instance, $E_H-E_L$ has a mean difference of $4.03$ in extraversion and a mean difference of $1.54$ in conscientiousness. Such correlation between different factors can be observed in similar work~\cite{smith2017understanding, sonlu2021conversational}. We ran the Kaiser-Meyer-Olkin test on our data to see if this correlation was too high; however, the resulting value of $.499$ indicates that it is unsuitable for factor analysis, and hence, the orthogonality of the samples is well captured.

\begin{table}[htbp]
\centering
\scriptsize
\caption{Pairwise comparison of the selected high ($X_H$), neutral ($X_N$), and low ($X_L$) samples, where $X \in \{O,C,E,A,N\}$. The rows show the personality factor that is compared; we report the mean difference and Tukey HSD adjusted p values for each sample pair. Bold values show $p < 0.05$.}
\setlength{\tabcolsep}{3pt}
\begin{tabular}{|c|lrr|lrr|lrr|lrr|lrr|}
\hline
\multicolumn{1}{|c|}{\textbf{\begin{tabular}[c]{@{}c@{}}Factor\end{tabular}}} & \multicolumn{1}{c}{\textbf{\begin{tabular}[c]{@{}c@{}}Sample\\ Pair\end{tabular}}} & \multicolumn{1}{c}{\textbf{\begin{tabular}[c]{@{}c@{}}Mean\\ Diff.\end{tabular}}} & \multicolumn{1}{c|}{\textbf{\begin{tabular}[c]{@{}c@{}}Adj.\\ P\end{tabular}}} & \multicolumn{1}{c}{\textbf{\begin{tabular}[c]{@{}c@{}}Sample\\ Pair\end{tabular}}} & \multicolumn{1}{c}{\textbf{\begin{tabular}[c]{@{}c@{}}Mean\\ Diff.\end{tabular}}} & \multicolumn{1}{c|}{\textbf{\begin{tabular}[c]{@{}c@{}}Adj.\\ P\end{tabular}}} & \multicolumn{1}{c}{\textbf{\begin{tabular}[c]{@{}c@{}}Sample\\ Pair\end{tabular}}} & \multicolumn{1}{c}{\textbf{\begin{tabular}[c]{@{}c@{}}Mean\\ Diff.\end{tabular}}} & \multicolumn{1}{c|}{\textbf{\begin{tabular}[c]{@{}c@{}}Adj.\\ P\end{tabular}}} & \multicolumn{1}{c}{\textbf{\begin{tabular}[c]{@{}c@{}}Sample\\ Pair\end{tabular}}} & \multicolumn{1}{c}{\textbf{\begin{tabular}[c]{@{}c@{}}Mean\\ Diff.\end{tabular}}} & \multicolumn{1}{c|}{\textbf{\begin{tabular}[c]{@{}c@{}}Adj.\\ P\end{tabular}}} & \multicolumn{1}{c}{\textbf{\begin{tabular}[c]{@{}c@{}}Sample\\ Pair\end{tabular}}} & \multicolumn{1}{c}{\textbf{\begin{tabular}[c]{@{}c@{}}Mean\\ Diff.\end{tabular}}} & \multicolumn{1}{c|}{\textbf{\begin{tabular}[c]{@{}c@{}}Adj.\\ P\end{tabular}}} \\ \hline
\multirow{3}{*}{\textbf{O}} & \textbf{$O_{H}-O_{L}$} & \textbf{2.412} & \textbf{\textless .001} & $C_{H}-C_{L}$ & -0.494 & .999 & $E_{H}-E_{L}$ & 1.130 & .573 & $A_{H}-A_{L}$ & -0.165 & .999 & $N_{H}-N_{L}$ & -0.179 & .999 \\
 & $O_{H}-O_{N}$ & 1.107 & .549 & $C_{H}-C_{N}$ & 0.149 & .999 & $E_{H}-E_{N}$ & 0.639 & .999 & $A_{H}-A_{N}$ & 0.308 & .999 & $N_{H}-N_{N}$ & 1.019 & .795 \\
 & $O_{N}-O_{L}$ & 1.305 & .191 & $C_{N}-C_{L}$ & -0.643 & .999 & $E_{N}-E_{L}$ & 0.491 & .999 & $A_{N}-A_{L}$ & -0.472 & .999 & $N_{N}-N_{L}$ & -1.198 & .387 \\ \hline
\multirow{3}{*}{\textbf{C}} & $O_{H}-O_{L}$ & -0.223 & .999 & \textbf{$C_{H}-C_{L}$} & \textbf{2.077} & \textbf{\textless .001} & \textbf{$E_{H}-E_{L}$} & \textbf{1.540} & \textbf{.006} & $A_{H}-A_{L}$ & 1.030 & .557 & $N_{H}-N_{L}$ & -1.286 & .074 \\
 & $O_{H}-O_{N}$ & -0.107 & .999 & $C_{H}-C_{N}$ & 0.792 & .985 & $E_{H}-E_{N}$ & 0.385 & .999 & $A_{H}-A_{N}$ & 0.001 & .999 & \textbf{$N_{H}-N_{N}$} & \textbf{-1.615} & \textbf{.002} \\
 & $O_{N}-O_{L}$ & -0.115 & .999 & $C_{N}-C_{L}$ & 1.286 & .074 & $E_{N}-E_{L}$ & 1.155 & .220 & $A_{N}-A_{L}$ & 1.030 & .557 & $N_{N}-N_{L}$ & 0.330 & .999 \\ \hline
\multirow{3}{*}{\textbf{E}} & \textbf{$O_{H}-O_{L}$} & \textbf{1.901} & \textbf{.001} & $C_{H}-C_{L}$ & -0.167 & .999 & \textbf{$E_{H}-E_{L}$} & \textbf{4.030} & \textbf{\textless .001} & $A_{H}-A_{L}$ & 0.134 & .999 & $N_{H}-N_{L}$ & -0.786 & .997 \\
 & \textbf{$O_{H}-O_{N}$} & \textbf{1.893} & \textbf{.001} & $C_{H}-C_{N}$ & -0.381 & .999 & \textbf{$E_{H}-E_{N}$} & \textbf{1.694} & \textbf{.009} & \textbf{$A_{H}-A_{N}$} & \textbf{1.577} & \textbf{.040} & \textbf{$N_{H}-N_{N}$} & \textbf{-1.635} & \textbf{.017} \\
 & $O_{N}-O_{L}$ & 0.008 & .999 & $C_{N}-C_{L}$ & 0.214 & .999 & \textbf{$E_{N}-E_{L}$} & \textbf{2.336} & \textbf{\textless .001} & $A_{N}-A_{L}$ & -1.443 & .093 & $N_{N}-N_{L}$ & 0.849 & .991 \\ \hline
\multirow{3}{*}{\textbf{A}} & $O_{H}-O_{L}$ & 1.459 & .096 & \textbf{$C_{H}-C_{L}$} & \textbf{1.708} & \textbf{.013} & $E_{H}-E_{L}$ & -0.367 & .999 & \textbf{$A_{H}-A_{L}$} & \textbf{3.546} & \textbf{\textless .001} & $N_{H}-N_{L}$ & 0.321 & .999 \\
 & $O_{H}-O_{N}$ & -0.429 & .999 & $C_{H}-C_{N}$ & 1.530 & .069 & $E_{H}-E_{N}$ & -0.726 & .999 & \textbf{$A_{H}-A_{N}$} & \textbf{1.615} & \textbf{.030} & $N_{H}-N_{N}$ & 1.530 & .051 \\
 & \textbf{$O_{N}-O_{L}$} & \textbf{1.887} & \textbf{.001} & $C_{N}-C_{L}$ & 0.179 & .999 & $E_{N}-E_{L}$ & 0.358 & .999 & \textbf{$A_{N}-A_{L}$} & \textbf{1.931} & \textbf{.001} & $N_{N}-N_{L}$ & -1.209 & .493 \\ \hline
\multirow{3}{*}{\textbf{N}} & $O_{H}-O_{L}$ & 0.706 & .999 & $C_{H}-C_{L}$ & -0.827 & .999 & $E_{H}-E_{L}$ & -1.357 & .367 & $A_{H}-A_{L}$ & -1.464 & .162 & \textbf{$N_{H}-N_{L}$} & \textbf{3.000} & \textbf{\textless .001} \\
 & $O_{H}-O_{N}$ & 0.786 & .999 & $C_{H}-C_{N}$ & -0.970 & .980 & $E_{H}-E_{N}$ & 0.828 & .999 & $A_{H}-A_{N}$ & 0.885 & .996 & \textbf{$N_{H}-N_{N}$} & \textbf{1.742} & \textbf{.017} \\
 & $O_{N}-O_{L}$ & -0.080 & .999 & $C_{N}-C_{L}$ & 0.143 & .999 & \textbf{$E_{N}-E_{L}$} & \textbf{-2.185} & \textbf{\textless .001} & \textbf{$A_{N}-A_{L}$} & \textbf{-2.349} & \textbf{\textless .001} & $N_{N}-N_{L}$ & 1.258 & .544 \\ \hline
\end{tabular}
\label{tab:tukey-pairwise-study-1}
\end{table}

\begin{figure*}[htbp]
  \centering
  \includegraphics[width=0.95\linewidth]{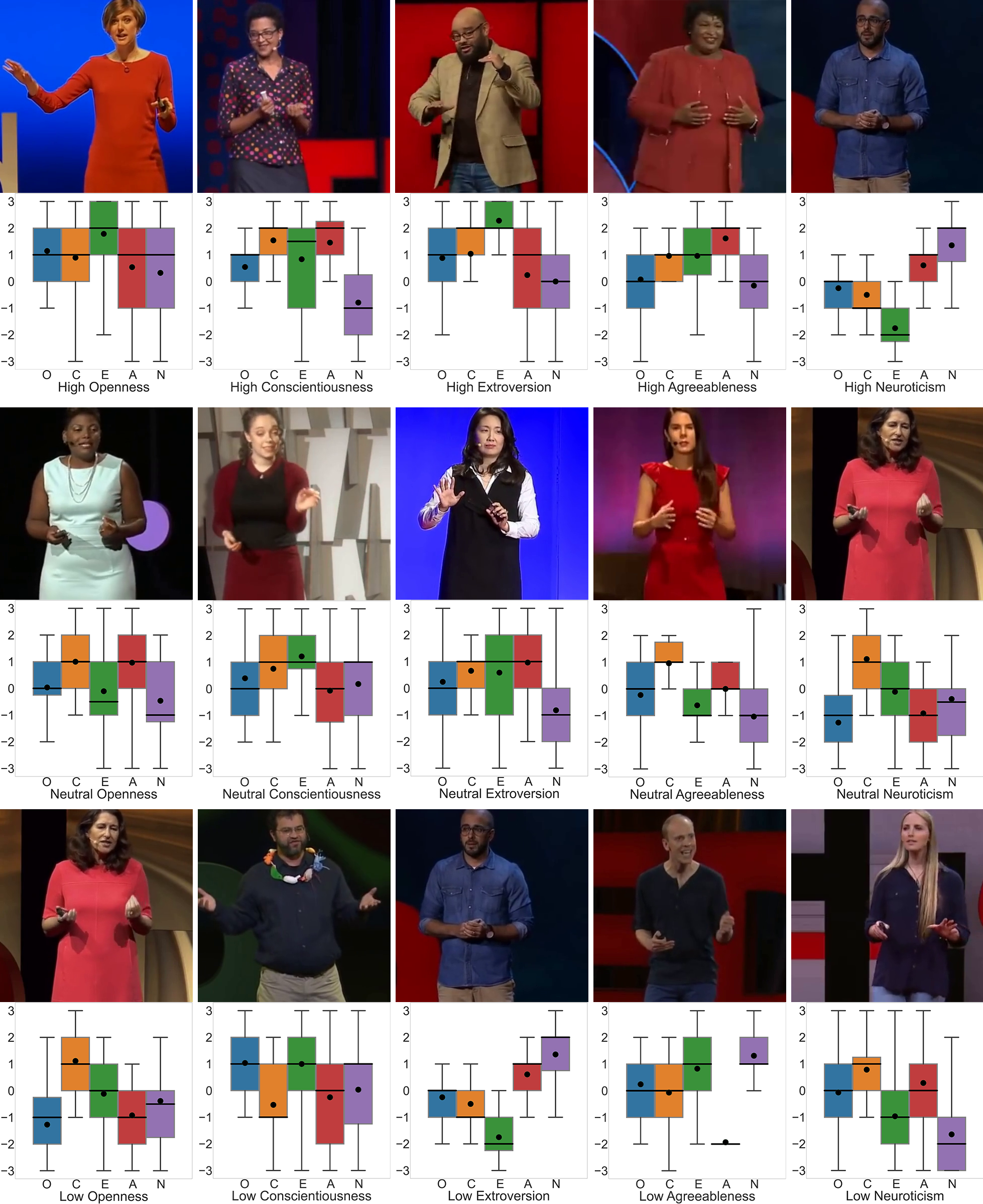}
  \caption{Selected samples representing high, low, and neutral traits for the corresponding personality factors. Each box plot summarizes the personality scores of the sample below; black lines depict the median, and the black dots show the mean.}
  \label{fig:all-ocean}
\end{figure*}

We illustrate the selected high, neutral, and low samples per factor in Figure~\ref{fig:all-ocean} with their personality block diagrams. Note that the samples of low extraversion - high neuroticism and low openness - neutral neuroticism are the same; this does not cause a problem for the second study as we focus on each factor individually, and there is no case where samples representing different factors are combined. Although we only report the personality ratings and pairwise comparisons for the selected samples in the article, the complete data and comparisons are available on our public repository for further analysis.

\subsection{Altering Samples for the Second Study}

We use TPS to alter the selected samples representing high, neutral, and low personality traits. For this task, we utilize different videos as TPS's source and driving inputs to output the video pairs introduced in Table~\ref{tab:cases}. We form these output pairs for each personality factor. We aim to cover as many of the possible combinations while keeping the sample size manageable. We compare these altered samples to determine which combination represents the high traits better in our second study.

\begin{table}[htbp]
\centering
\caption{Motion and appearance combinations used in the second study.}
\begin{tabular}{|c|cc|cc|}
\hline
\multicolumn{1}{|l|}{\textbf{}} & \multicolumn{2}{c|}{\textbf{Output A}} & \multicolumn{2}{c|}{\textbf{Output B}} \\ \hline
\textbf{Case} & \multicolumn{1}{c|}{\textbf{Source}} & \textbf{Driving} & \multicolumn{1}{c|}{\textbf{Source}} & \textbf{Driving} \\ \hline
1 & \multicolumn{1}{c|}{Neutral} & High & \multicolumn{1}{c|}{Neutral} & Low \\ \hline
2 & \multicolumn{1}{c|}{High} & Neutral & \multicolumn{1}{c|}{Low} & Neutral \\ \hline
3 & \multicolumn{1}{c|}{Low} & High & \multicolumn{1}{c|}{High} & Low \\ \hline
4 & \multicolumn{1}{c|}{High} & High & \multicolumn{1}{c|}{High} & Low \\ \hline
5 & \multicolumn{1}{c|}{Low} & High & \multicolumn{1}{c|}{Low} & Low \\ \hline
\end{tabular}
\label{tab:cases}
\end{table}

Three cases measure the impact of opposing trait motions on the same source image. We use the samples selected for neutral, high, and low traits as the source input in Cases 1, 4, and 5, respectively; we determine if the source image affects perception when appearance is the same and motions are opposing. Case 2 investigates the effect of appearance where we use the neutral sample's motion as the driving input, and opposing traits are used as source inputs; we only use the neutral motion and omit the combinations that use high and low samples as the driving input on the opposing sources to keep the focus on the appearance. As the motions are the same for these combinations, we expect a similar motion transfer quality; in contrast, as Cases 1, 4, and 5 cover different source images, the performance of TPS can vary, which could influence perception. We discuss these possible artifacts due to the mismatch between source and driving inputs in the Limitations section.

In each comparison, we use high and low samples as the source or driving input to keep the stimuli more practical; we do not compare high and neutral driving inputs on the same source or any such combination. Case 3 analyzes whether driving or source input is more dominant when they utilize opposing samples. We find out which combination expresses the high trait better: Low trait appearance with high trait motion or high trait appearance with low trait motion. For each personality factor, we have five cases resulting in 25 output pairs.

\subsection{Motion Transfer}

We utilize the pre-trained version of TPS, which takes an RGB source image and a driving video of 384 $\times$ 384 resolution for motion transfer; the motion of the driving video is transferred to the actor in the source image. The synthesized video has the exact frame count as the driving video. As our samples are videos, we need to choose a video frame to be used as the source image, and the pose in the selected frame determines the quality of the output. In cases where the actor in the source image is blurry or has an abnormal pose, the output suffers.

We follow the recommended alignment method in the original TPS implementation to prevent low-quality results. The authors suggest finding an aligning frame in the driving video with a pose similar to the source image. The alignment is calculated by determining the face landmarks and calculating their 2D convex hull. The landmarks are centered and normalized according to the square root of the area of the convex hull. We take the frame that minimizes the squared error between source image landmarks as the representative frame. The motion transfer is initialized from this frame in the driving video and processed in both directions: toward the beginning and the end. The final videos are stitched accordingly, and the final output is obtained.

\begin{figure}[htbp]
  \centering
  \includegraphics[width=0.5\linewidth]{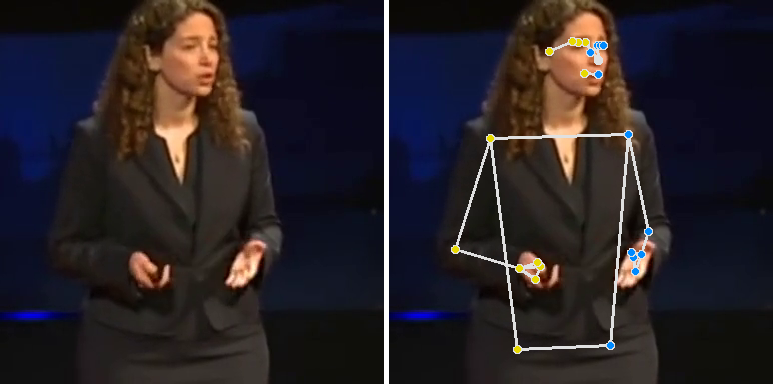}
  \caption{Example representative poses. Images on the right show the landmarks of the representative frames on the left.}
  \label{example_representatives}
\end{figure}

\begin{figure}[htbp]
  \centering
  \includegraphics[width=0.5\linewidth]{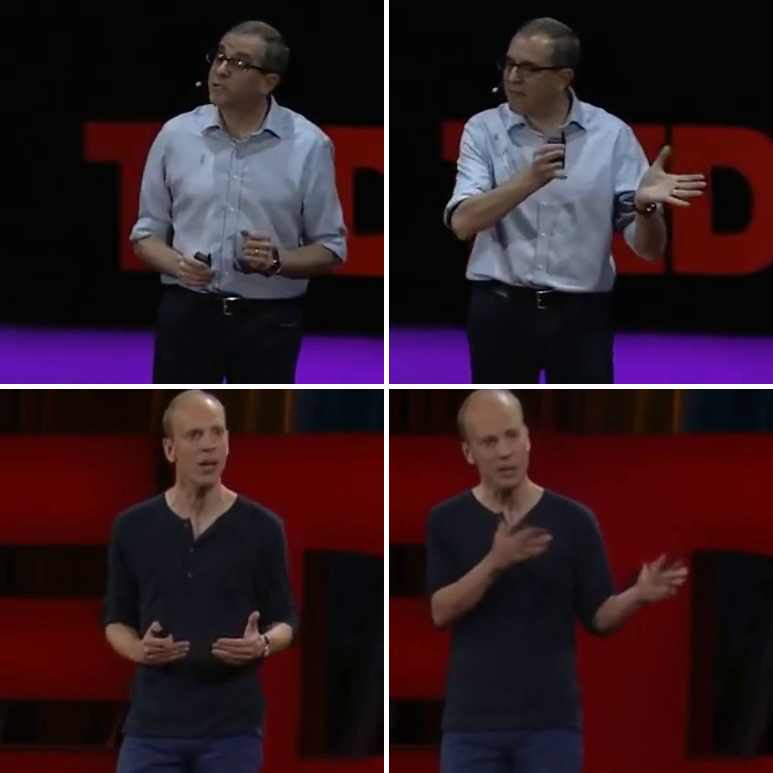}
  \caption{Sample motion transfer; the left column shows the representative frames from driving (top) and source (bottom) videos; the top right image shows the driving video and the bottom right image shows the transferred output.}
  \label{example_transformation}
\end{figure}

Since calculating the representative frame for each pair of videos would be time-consuming, we pre-process all videos using a standard idle pose. We first calculate 3D landmarks of the human body using MediaPipe Pose Landmarker~\cite{mediapipe2023} and calculate the 3D convex hull. The frame in each video closest to the idle pose is taken as the representative frame of that video.  An example representative frame and its extracted pose can be seen in Figure~\ref{example_representatives}. During motion transfer, the representative frame of the source video is taken as the source image. The stitching logic in the original work mentioned above is applied as is. An example motion transfer between a source image and a driving video can be seen in Figure~\ref{example_transformation}. Notice the slight deformation of the face in the source image as they face different directions. Such deformations can reduce the quality of the outputs, influencing the perception of desired traits.

\subsection{Second User Study}

The second study compares the altered output pairs introduced in Table~\ref{tab:cases}. In each task, we play the matched output pairs simultaneously. Participants are free to repeat, pause, and rewind the videos. The order and left-right placement of the outputs are randomized. Our first question asks, ``Which video contains a more \textbf{extraverted, enthusiastic} and less \textbf{reserved, quiet} looking person?''. We replace the keywords written in bold using the traits from Table~\ref{tab:opposite-traits} based on the personality factor we measure. The second question asks, ``Which video looks more realistic?''. For each question, the participant selects one of the buttons indicating \emph{Left}, \emph{Equal}, and \emph{Right}. We require watching the videos at least once before submitting. At the beginning of the study, we show a text briefly explaining the Five-Factor personality traits to the participants. We inform the participants that the samples are modified using software and define a realistic-looking sample as one with no noticeable marks of the modification. In this case, a less realistic sample has artifacts that reveal the modifications on the video.

We map the users' answers to the integers 0, 1, and 2 for choosing Output A, Equal, or Output B, respectively, to be used as the dependent variable. We compare output pairs of TPS following the combinations of five cases. We only focus on the personality factor that input samples represent. For example, in Case 1, we use $E_N$ (source) and $E_H$ (driving) to produce output $A_{E1}$ and $E_N$ (source) and $E_L$ (driving) to produce output $B_{E1}$. Outputs $A_{E1}$ and $B_{E1}$ are compared based on their performance in representing high extraversion. We have the output pairs of $A_{Xi}$ and $B_{Xi}$, where $X \in {O, C, E, A, N}$ representing personality and $i \in {1, 2, 3, 4, 5}$ representing the different cases.

We report ANOVA statistics of our second study in Table~\ref{tab:anova-study-2}, using two groupings: The first one separates the samples into different case groups ($i$) and uses personality ($X$) as the independent variable; the second one separates the samples into different personality groups and uses different cases as the independent variable. Separating into case groups helps analyze the performance of each approach in expressing distinct factors; for example, we observe that Cases 3 and 4 are the most successful ones in distinguishing different factors; on the other hand, the influence of Case 2 is very similar for each factor. Separating into personality groups helps analyze the expressibility of each personality dimension; for example, we observe that our cases best distinguish extraversion while the smallest influence is on conscientiousness. The linearity and sphericity conditions of repeated measures ANOVA hold for both groupings.

\begin{table}[htbp]
\centering
\caption{ANOVA statistics for the second user study considering case (first five rows) and personality (last five rows) groups. Statistics marked with ** indicate $p < 0.001$, and * indicate $p < 0.05$.}
\begin{tabular}{clcccc}
\hline
\textbf{} & \multicolumn{1}{c}{\textbf{Source}} & \multicolumn{1}{c}{\textbf{SS}} & \multicolumn{1}{c}{\textbf{df}} & \multicolumn{1}{c}{\textbf{F}} & \multicolumn{1}{c}{\textbf{$\eta^{2}$}} \\ \hline
\multirow{2}{*}{1} & Between groups & 7.37 & 4 & 3.305* & 0.083 \\
 & Within groups & 80.86 & 145 & -- & -- \\ \hline
\multirow{2}{*}{2} & Between groups & 7.13 & 4 & 2.375 & 0.061 \\
 & Within groups & 108.86 & 145 & -- & -- \\ \hline
\multirow{2}{*}{3} & Between groups & 10.37 & 4 & 10.632** & 0.226 \\
 & Within groups & 35.36 & 145 & -- & -- \\ \hline
\multirow{2}{*}{4} & Between groups & 25.22 & 4 & 10.662** & 0.227 \\
 & Within groups & 85.76 & 145 & -- & -- \\ \hline
\multirow{2}{*}{5} & Between groups & 17.10 & 4 & 6.824** & 0.158 \\
 & Within groups & 90.86 & 145 & -- & -- \\ \hline
\multirow{2}{*}{O} & Between groups & 20.49 & 4 & 9.019** & 0.199 \\
 & Within groups & 82.36 & 145 & -- & -- \\ \hline
\multirow{2}{*}{C} & Between groups & 8.24 & 4 & 3.732** & 0.093 \\
 & Within groups & 80.03 & 145 & -- & -- \\ \hline
\multirow{2}{*}{E} & Between groups & 39.16 & 4 & 17.518** & 0.325 \\
 & Within groups & 81.03 & 145 & -- & -- \\ \hline
\multirow{2}{*}{A} & Between groups & 21.37 & 4 & 10.167** & 0.219 \\
 & Within groups & 76.20 & 145 & -- & -- \\ \hline
\multirow{2}{*}{N} & Between groups & 28.04 & 4 & 12.380** & 0.254 \\
 & Within groups & 82.10 & 145 & -- & -- \\ \hline
\end{tabular}
\label{tab:anova-study-2}
\end{table}

We further inspect pairwise differences with Tukey~HSD; Table~\ref{tab:pair-study-2} shows the mean differences of personality factors per case and the mean differences of cases per personality factor. We observe Case 3 having the highest mean differences between different factors. In Case 3, Output A utilizes $X_H$ for the driving input, and Output B uses the driving input $X_L$. For extraversion, $A_{E3}$ is preferred for expressing the high traits, showing the importance of motion for this factor. In the same case, Output B is preferred for agreeableness and neuroticism, suggesting these factors are more related to appearance. In Case 2, where we utilize the same driving motion, agreeableness and neuroticism are easily distinguished due to the high trait source input. Openness and conscientiousness are barely realized in Cases 2 and 3.

\begin{table*}[htbp]
\centering
\scriptsize
\caption{Pairwise comparisons from the second study and the corresponding ANOVA statistics. We perform ANOVA on different cases where personality factors are independent variables (left half, pairs from $\{O, C, E, A, N\}$) and on different personality factors where cases are independent variables (right half, pairs from $\{1,2,3,4,5\}$. The dependent variable is the participants' choices mapped to integers 0, 1, and 2, corresponding to choosing Output A, Equal, and Output B, respectively. We report each pair's mean difference and Tukey HSD adjusted p values. Bold values show $p < 0.05$.}
\setlength{\tabcolsep}{2.9pt}
\begin{tabular}{|c|rr|rr|rr|rr|rr|c|rr|rr|rr|rr|rr|}
\hline
\textbf{Grp.} & \multicolumn{2}{c|}{\textbf{Case 1}} & \multicolumn{2}{c|}{\textbf{Case 2}} & \multicolumn{2}{c|}{\textbf{Case 3}} & \multicolumn{2}{c|}{\textbf{Case 4}} & \multicolumn{2}{c|}{\textbf{Case 5}} & \textbf{Grp.} & \multicolumn{2}{c|}{\textbf{O}} & \multicolumn{2}{c|}{\textbf{C}} & \multicolumn{2}{c|}{\textbf{E}} & \multicolumn{2}{c|}{\textbf{A}} & \multicolumn{2}{c|}{\textbf{N}} \\ \hline
\textbf{Pair} & \multicolumn{1}{c}{\textbf{\begin{tabular}[c]{@{}c@{}}M.\\ Diff.\end{tabular}}} & \multicolumn{1}{c|}{\textbf{\begin{tabular}[c]{@{}c@{}}Adj.\\ P\end{tabular}}} & \multicolumn{1}{c}{\textbf{\begin{tabular}[c]{@{}c@{}}M.\\ Diff.\end{tabular}}} & \multicolumn{1}{c|}{\textbf{\begin{tabular}[c]{@{}c@{}}Adj.\\ P\end{tabular}}} & \multicolumn{1}{c}{\textbf{\begin{tabular}[c]{@{}c@{}}M.\\ Diff.\end{tabular}}} & \multicolumn{1}{c|}{\textbf{\begin{tabular}[c]{@{}c@{}}Adj.\\ P\end{tabular}}} & \multicolumn{1}{c}{\textbf{\begin{tabular}[c]{@{}c@{}}M.\\ Diff.\end{tabular}}} & \multicolumn{1}{c|}{\textbf{\begin{tabular}[c]{@{}c@{}}Adj.\\ P\end{tabular}}} & \multicolumn{1}{c}{\textbf{\begin{tabular}[c]{@{}c@{}}M.\\ Diff.\end{tabular}}} & \multicolumn{1}{c|}{\textbf{\begin{tabular}[c]{@{}c@{}}Adj.\\ P\end{tabular}}} & \textbf{Pair} & \multicolumn{1}{c}{\textbf{\begin{tabular}[c]{@{}c@{}}M.\\ Diff.\end{tabular}}} & \multicolumn{1}{c|}{\textbf{\begin{tabular}[c]{@{}c@{}}Adj.\\ P\end{tabular}}} & \multicolumn{1}{c}{\textbf{\begin{tabular}[c]{@{}c@{}}M.\\ Diff.\end{tabular}}} & \multicolumn{1}{c|}{\textbf{\begin{tabular}[c]{@{}c@{}}Adj.\\ P\end{tabular}}} & \multicolumn{1}{c}{\textbf{\begin{tabular}[c]{@{}c@{}}M.\\ Diff.\end{tabular}}} & \multicolumn{1}{c|}{\textbf{\begin{tabular}[c]{@{}c@{}}Adj.\\ P\end{tabular}}} & \multicolumn{1}{c}{\textbf{\begin{tabular}[c]{@{}c@{}}M.\\ Diff.\end{tabular}}} & \multicolumn{1}{c|}{\textbf{\begin{tabular}[c]{@{}c@{}}Adj.\\ P\end{tabular}}} & \multicolumn{1}{c}{\textbf{\begin{tabular}[c]{@{}c@{}}Mean\\ Diff.\end{tabular}}} & \multicolumn{1}{c|}{\textbf{\begin{tabular}[c]{@{}c@{}}Adj.\\ P\end{tabular}}} \\ \hline
\textbf{C-O} & 0.46 & .121 & 0.01 & .999 & 0.30 & .529 & 0.23 & .724 & 0.40 & .243 & \textbf{2-1} & 0.40 & .236 & -0.06 & .998 & \textbf{0.56} & \textbf{.001} & \textbf{-0.86} & \textbf{.001} & -0.30 & .585 \\ \hline
\textbf{E-O} & -0.36 & .330 & -0.20 & .835 & \textbf{-0.83} & \textbf{.001} & -0.33 & .388 & \textbf{-0.76} & \textbf{.001} & \textbf{3-1} & 0.43 & .168 & 0.26 & .756 & -0.03 & .999 & 0.33 & .450 & \textbf{0.66} & \textbf{.011} \\ \hline
\textbf{A-O} & \textbf{0.70} & \textbf{.004} & \textbf{-0.56} & \textbf{.029} & \textbf{0.60} & \textbf{.018} & 0.36 & .291 & 0.33 & .427 & \textbf{4-1} & -0.13 & .958 & -0.36 & .475 & -0.10 & .934 & -0.46 & .135 & 0.43 & .217 \\ \hline
\textbf{N-O} & 0.23 & .751 & -0.46 & .112 & 0.46 & .116 & \textbf{0.80} & \textbf{.001} & 0.30 & .535 & \textbf{5-1} & 0.23 & .745 & 0.16 & .945 & -0.16 & .687 & -0.13 & .962 & 0.30 & .585 \\ \hline
\textbf{E-C} & \textbf{-0.83} & \textbf{.001} & -0.20 & .835 & \textbf{-1.13} & \textbf{.001} & \textbf{-0.56} & \textbf{.024} & \textbf{-1.16} & \textbf{.001} & \textbf{3-2} & 0.03 & .999 & 0.33 & .570 & \textbf{-0.60} & \textbf{.001} & \textbf{1.20} & \textbf{.001} & \textbf{0.96} & \textbf{.001} \\ \hline
\textbf{A-C} & 0.23 & .751 & \textbf{-0.56} & \textbf{.029} & 0.30 & .529 & 0.13 & .953 & -0.06 & .997 & \textbf{4-2} & \textbf{-0.53} & \textbf{.049} & -0.30 & .666 & \textbf{-0.66} & \textbf{.001} & 0.40 & .264 & \textbf{0.73} & \textbf{.004} \\ \hline
\textbf{N-C} & -0.23 & .751 & -0.46 & .112 & 0.16 & .909 & \textbf{0.56} & \textbf{.024} & -0.10 & .985 & \textbf{5-2} & -0.16 & .909 & 0.23 & .834 & \textbf{-0.73} & \textbf{.001} & \textbf{0.73} & \textbf{.002} & \textbf{0.60} & \textbf{.031} \\ \hline
\textbf{A-E} & \textbf{1.06} & \textbf{.001} & -0.36 & .315 & \textbf{1.43} & \textbf{.001} & \textbf{0.70} & \textbf{.002} & \textbf{1.10} & \textbf{.001} & \textbf{4-3} & \textbf{-0.56} & \textbf{.030} & \textbf{-0.63} & \textbf{.041} & -0.06 & .984 & \textbf{-0.80} & \textbf{.001} & -0.23 & .784 \\ \hline
\textbf{N-E} & \textbf{0.60} & \textbf{.020} & -0.26 & .635 & \textbf{1.30} & \textbf{.001} & \textbf{1.13} & \textbf{.001} & \textbf{1.06} & \textbf{.001} & \textbf{5-3} & -0.20 & .837 & -0.10 & .991 & -0.13 & .833 & -0.46 & .135 & -0.36 & .381 \\ \hline
\textbf{N-A} & -0.46 & .121 & 0.10 & .985 & -0.13 & .958 & 0.43 & .146 & -0.03 & .999 & \textbf{5-4} & 0.36 & .321 & 0.53 & .125 & -0.06 & .984 & 0.33 & .450 & -0.13 & .966 \\ \hline
\end{tabular}
\label{tab:pair-study-2}
\end{table*}

In Cases 1, 4, and 5, we expect the means to be close to 0, as in Output A, we utilize the high trait sample as driving input, and the source inputs of both outputs are the same. Extraversion behaves as expected; this is not true for other factors. For example, using low trait source input for conscientiousness and agreeableness factors causes Output B to be slightly confused for high traits, causing their means to get close to 1; however, the difference between Cases~1,~4,~and~5 is not significant.

\begin{figure*}[htbp]
  \centering
  \includegraphics[width=0.95\linewidth]{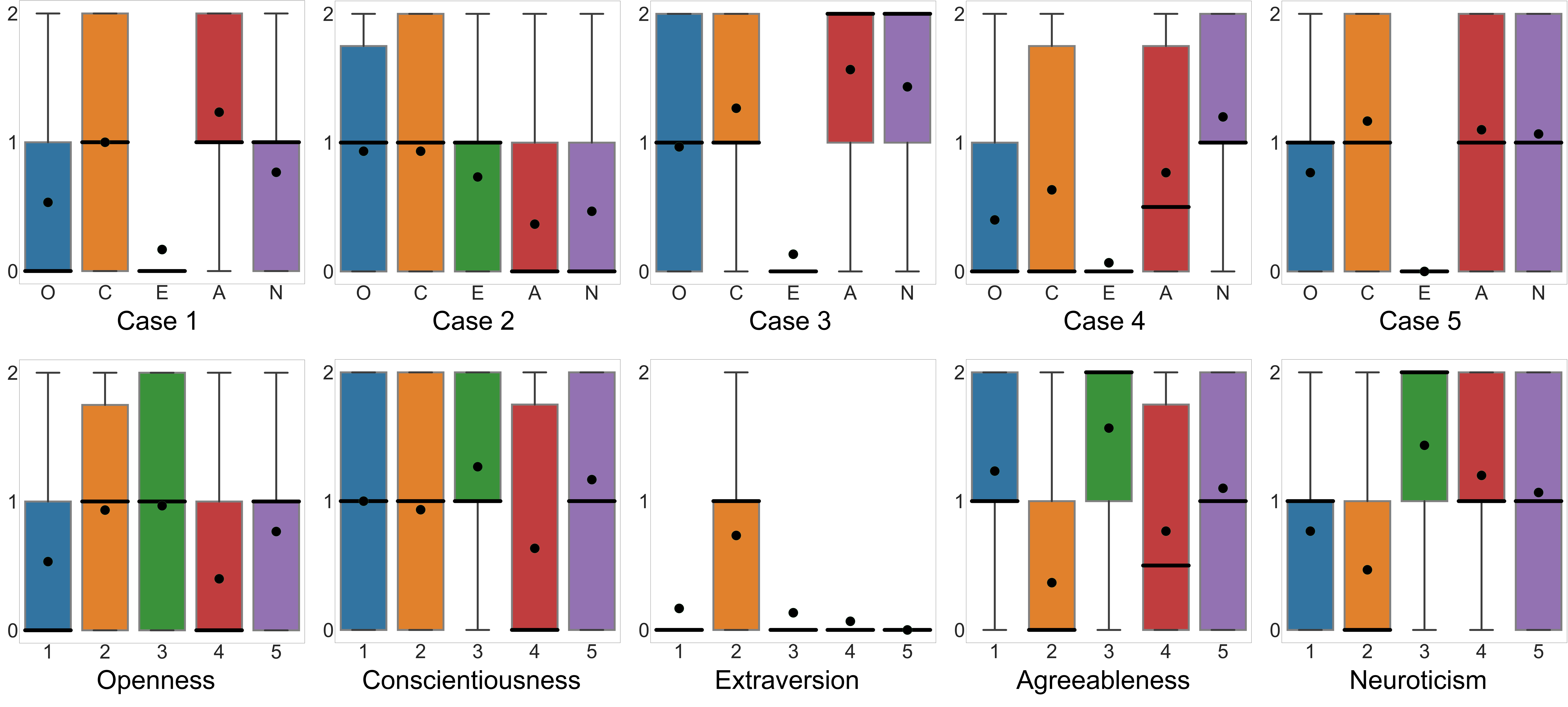}
  \caption{Box plots for the second study where participants' choice for Output A, Equal, and Output B are mapped to integers 0, 1, and 2, respectively. Black lines depict the median, and the black dots show the mean. Means close to 1 show that the corresponding combination does not help distinguish the high and low traits of the related personality. The first row examines the factors grouped by case, and the second row examines the cases grouped by personality.}
  \label{fig:means-second}
\end{figure*}

Cases 1 and 2 significantly differ in extraversion and agreeableness; extraversion is better distinguished when the motions differ, and agreeableness is better when appearances differ. The results of the different cases agree that the driving input can express extraversion better, and the source input represents agreeableness and neuroticism better. Cases 1, 2, and 5 are not successful in creating a distinction in conscientiousness; this can be due to this factor being hard to observe in short sequences. Additionally, our original samples had a lower variance in conscientiousness, which can be the reason for this reduced effect. Similarly, openness is not distinguished sufficiently for Cases 2, 3, and 5. Figure~\ref{fig:means-second} shows box plots for the same groupings.

Extraversion is the most distinguished factor as motion cues represent perceived extraversion well, and all cases except for Case 2 utilize opposing motions; utilizing the same motion, distinguishing extraversion becomes challenging. Agreeableness and neuroticism are better distinguished through appearance cues. Openness is best expressed through motion when neutral or high appearance is used; however, using the low trait sample as the source input diminishes openness perception. Conscientiousness is best observed in Case 3 through its driving input and is hardly distinguished in any other case.

We perform Pearson correlation analysis between the answers to the first and second questions to determine if the output's realism is related to personality perception. Although realism is essential in expressing personality~\cite{ZibrekEtAl19}, we do not find a consistent relationship between the participants' personality and realism choices. This observation is likely because, in our case, realism is only harmed by TPS's artifacts when the source and driving inputs are too different in terms of their pose. Unlike a difference in rendering quality or style, the artifacts do not alter the outputs dramatically; the figures' movement and appearance are still readable, with occasional deformations and blending problems. People can read human motion accurately even with minimal information such as point light data~\cite{blake2007perception}. As long as the hands and elbows of the figure are visible, which is the case in our outputs, we do not expect confusion for the motion. Similarly, the deformations do not cause a change in the stylistic elements of the character; thus, we do not expect them to alter the character's image in the participants' minds.

Similar to our personality analysis, we map participants' answers to the integers 0 (Output A), 1 (Equal), and 2 (Output B) for the question asking which output looks realistic compared to others. We report the Pearson correlation coefficients between the participants' answers to realism and personality choices in Table~\ref{tab:pearson}. A linear relationship is primarily visible in Case 3, where opposing trait samples are used as driving and source inputs. The correlation is weak for conscientiousness and extraversion and moderate for agreeableness and neuroticism, suggesting some participants may have perceived the personality of the output concerning their realism. On the other hand, such a relationship is not visible in other cases. Especially in Cases 4 and 5, we see a clear preference for the outputs of matching trait inputs having high realism; Case 4 offers the optimal combination for Output A (High-High), and Case 5 offers the optimal combination for Output B (Low-Low); these outputs received a relatively high preference for realism, yet they do not have a significant correlation to the answers to the personality question. Consequently, we believe the influence of the occasional artifacts in outputs does not have a visible effect on the experiments.

\begin{table}[htbp]
\centering
\caption{Pearson correlation coefficients (Cor.) between the participants' answers to realism and personality questions and their P-values. Bold values show $p < 0.05$. We assume $Cor. < 0.4$ as weak and $0.4 < Cor. < 0.8$ as moderate.}
\setlength{\tabcolsep}{3pt}
\begin{tabular}{|c|c|r|r|r|r|r|}
\hline
\multicolumn{1}{|l|}{\textbf{Factor}} & \multicolumn{1}{l|}{\textbf{Case:}} & \multicolumn{1}{c|}{\textbf{1}} & \multicolumn{1}{c|}{\textbf{2}} & \multicolumn{1}{c|}{\textbf{3}} & \multicolumn{1}{c|}{\textbf{4}} & \multicolumn{1}{c|}{\textbf{5}} \\ \hline
\multirow{2}{*}{\textbf{O}} & \textbf{Cor.} & \textbf{.534} & .145 & -.093 & \textbf{.417} & .187 \\ \cline{2-7} 
 & \textbf{P} & \textbf{.002} & .444 & .625 & \textbf{.022} & .321 \\ \hline
\multirow{2}{*}{\textbf{C}} & \textbf{Cor.} & .338 & .323 & \textbf{.386} & \textbf{.436} & .270 \\ \cline{2-7} 
 & \textbf{P} & .068 & .082 & \textbf{.035} & \textbf{.016} & .149 \\ \hline
\multirow{2}{*}{\textbf{E}} & \textbf{Cor.} & .193 & .296 & \textbf{.378} & .240 & .122 \\ \cline{2-7} 
 & \textbf{P} & .306 & .112 & \textbf{.040} & .202 & .522 \\ \hline
\multirow{2}{*}{\textbf{A}} & \textbf{Cor.} & .339 & .340 & \textbf{.502} & .315 & .075 \\ \cline{2-7} 
 & \textbf{P} & .067 & .066 & \textbf{.005} & .090 & .695 \\ \hline
\multirow{2}{*}{\textbf{N}} & \textbf{Cor.} & .055 & -.119 & \textbf{-.515} & .066 & -.170 \\ \cline{2-7} 
 & \textbf{P} & .772 & .533 & \textbf{.004} & .727 & .369 \\ \hline
\end{tabular}
\label{tab:pearson}
\end{table}

\begin{figure*}[htbp]
  \centering
  \includegraphics[width=0.95\linewidth]{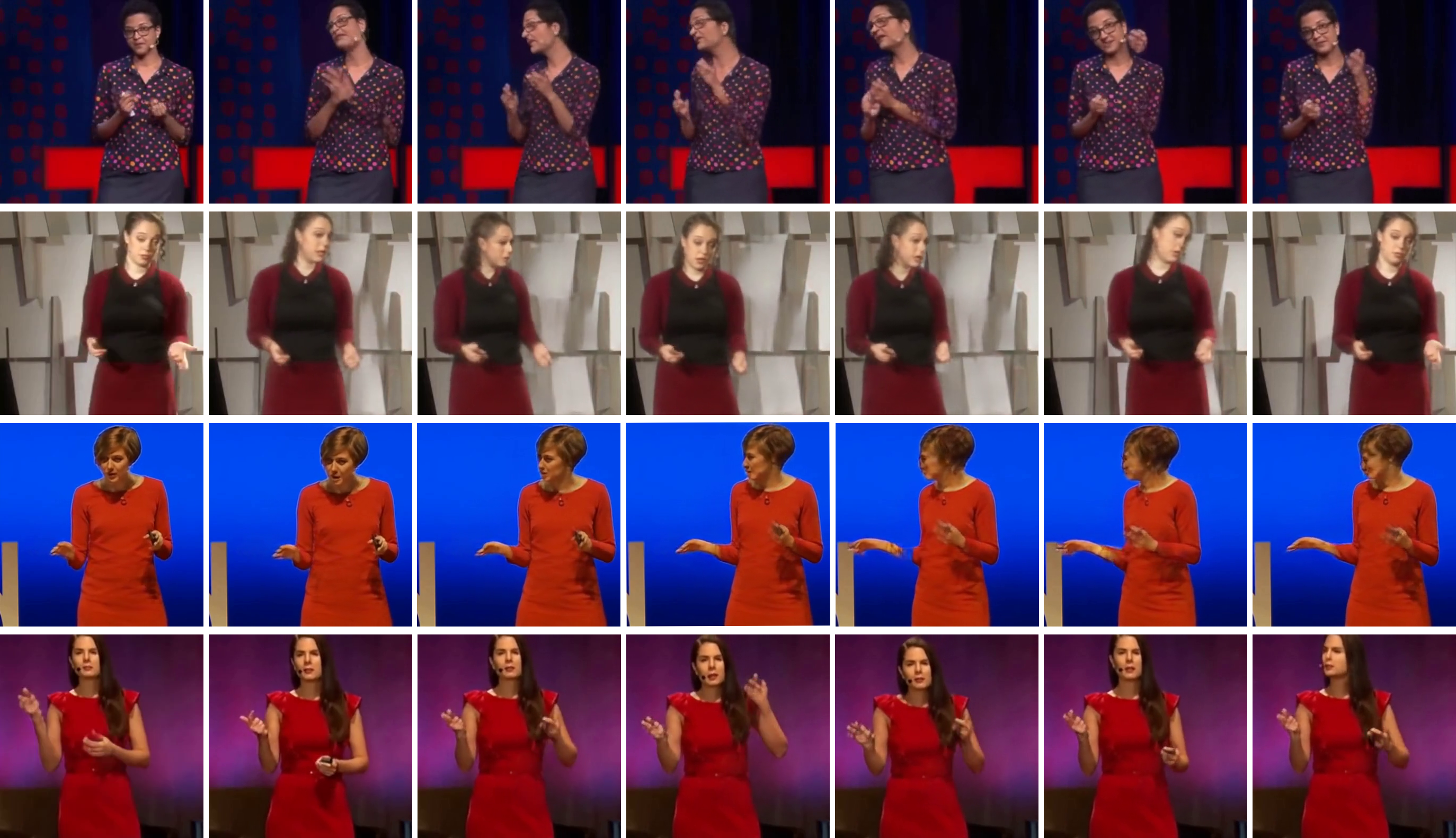}
  \caption{Consecutive frames of example outputs from TPS. Rows illustrate the frames that cover short segments of the outputs $A_{C2}$, $A_{C1}$, $A_{O2}$, $B_{A1}$, in order. The first output has artifacts where limb parts are mixed with the background, and the second output has artifacts in the background; the third output has a deformed face due to differences in facing directions. In contrast, the fourth output has no obvious artifacts.}
  \label{fig:artifacts}
\end{figure*}

We illustrate some of the outputs that contain occasional artifacts in Figure~\ref{fig:artifacts}. We show the consecutive frames of one-second-long output segments that contain artifacts. In the first output, the forearm of the figure is mixed with the cloth texture; this is likely due to TPS's confusion regarding the boundary of cloth and skin. The second output has a non-stationary background due to the figure changing position. The model fills in the background details, but the output can contain artifacts when the original background has high complexity; this is not prominent when the source input has a simple background. The third output contains artifacts due to a mismatch between the facing directions of the source and driving inputs; the source is facing forward with a slightly tilted head, and a part of the driving input requires the head to turn left. The same output also has a problem with the forearm of the figure blending into the background. In contrast, the last output has no obvious artifacts as its source contains a simple background, and its source and driving inputs have well-matching poses. Additionally, the driving input does not have dramatic pose changes, which also helps the output to have better quality.  Generally, the artifacts are due to TPS having less information about the source input; future work can improve output quality by supporting multiple inputs from different views to fill in the missing details.

\section{Discussion}
Our first study approves that participants' zero-acquaintance personality judgments generally agree. Despite minimal information about the individual, participants observe personality traits in short silent videos showing the subject's upper body. In this case, the participants watch the video until the end; the subject's movements, clothes, facial expressions, or the scene's background and lighting can all influence the perceived personality. Future work can dissect each element to analyze their influence individually; however, in this work, we assumed that apparent personality is mainly due to movements and looks.

Our second study analyzes TPS's performance in personality expression using the selected samples. Following our first research question, Cases 1 and 2 results show that TPS can alter personality perception by utilizing different driving and source inputs on neutral appearance and motion, respectively. Participants perceive the videos whose appearance or motion is altered by TPS differently; this suggests that applications can utilize TPS to alter personality perception of prerecorded videos. For example, an intelligent assistant expressing different personality traits based on the user may utilize TPS to create variation with less effort.

The output pairs we form using TPS focus on analyzing the influence of two main elements for personality expression: Appearance and motion. Case 1 analyzes the influence of motion by using the opposing driving inputs on the same source input. We observe that the perceived extraversion of the sample highly depends on its motion; the driving input also influences the perceived openness but with less confidence. Case 2 analyzes the influence of appearance using the opposing source inputs on the same driving input. The results suggest that perceived agreeableness and neuroticism highly depend on the subject's appearance. Following our second research question, the impact of motion and appearance on perceived personality is not the same; motion better reflects extraversion and openness, while appearance better reflects agreeableness and neuroticism. This phenomenon is best observed in our Case 3, where the motion and appearance of the output utilize opposing trait samples. The high extraversion motion with the low extraversion appearance is perceived as high extraversion, indicating the importance of motion for extraversion. On the other hand, the opposite is true for agreeableness; high agreeableness appearance with low agreeableness motion is perceived as high agreeableness. Cases 4 and 5 agree with the results of Cases 1 and 2.

We analyze the results of our second user study by grouping the samples according to case and personality. The case groups help us observe whether using different combinations influences the personality factors similarly. We observe that certain cases, such as 3 and 4, cause a higher variance between groups; they are more successful in representing distinct factors. On the other hand, Case 2 has less variance between groups, suggesting the participants perceived the different personality factors together when utilizing the appearance. Following our third research question, the influence of TPS varies for different factors; different cases control different factors dominantly, and the correlation between factors changes based on the case. Grouping by personality helps see the overall effect of TPS on personality perception; a significant difference can be observed for all personality factors using appearance or motion: the highest distinction is for extraversion, and the lowest is for conscientiousness.

TPS's performance may depend on how much the inputs express the corresponding traits. For example, transferring the motion of a slightly high extraversion sample to the same neutral source can be less effective than transferring the motion of a relatively higher extraversion sample. Future studies can utilize TPS inputs on a wider scale. Although TPS's output had occasional artifacts in certain cases, we did not observe a significant effect due to the realism or quality of the compared outputs. Advances in motion transfer models can enable future work to express certain factors better. For example, the model we used cannot transfer the mimics of the driving input in great detail; however, including such subtle details can improve the performance. Recent work in motion generation utilizes diffusion models to generate text-driven human animations~\cite{zhang2022motiondiffuse,yuan2023physdiff, tevet2023human}, and style-based motion~\cite{ghorbani2022zeroeggs,ao2023gesturediffuclip}. Future work can utilize motion generation models to control the stimuli more.

\section{Limitations}
Although we aim to analyze the implications of personality in altered videos that adopt certain looks and motions with a thorough study, the vastness of the field leads us to use certain simplifications. First, the typical usage of TIPI asks one question for each polarity of personality that is averaged with the opposite trait to measure the related factor more precisely. In our case, the sensitivity and psychometric properties of the instrument are harmed to shorten the time required for evaluating each sample. The average time to complete the first user study for a participant was 37 minutes.

The feedback from the participants indicated that they found the study long. Since the second user study involved comparisons between output pairs, we resorted to further simplifications. Instead of asking personality questions involving all five factors, we asked a single question focusing on the factor output pairs use. Since we aimed to capture the changes in one personality dimension in each case, this simplification did not harm our analysis; however, with more data captured, we could also analyze if the changes affected other factors simultaneously.

In the second study, we used the choices left, equal, and right instead of using a Likert scale for the participants' answers. While this causes a loss of precision, using a scale for a question that asks which sample expresses the high traits better could have caused confusion. Furthermore, we are interested in whether different input combinations of TPS work similarly for different personality factors rather than measuring how effective each is; consequently, this brief approach helps shorten the average study time without obstructing our analysis. The average time to complete the second study was 18 minutes.

Our participant count is relatively low for a perception study; we tried to keep the sample size low to compensate for this, which could cause the results not to generalize well in a bigger dataset. In particular, all the individuals used in the videos are TED presenters, which may cause them to have a bias in certain factors like conscientiousness and agreeableness. On the other hand, since we utilize short clips from the talks, the current motion and appearance become more influential on the apparent personality. We also compensate for the low participation count by recruiting participants from reliable sources. Although we ran the study online, all the participants were current university students and their family members. With the spread of paid online services for running user studies, the quality of the responses becomes questionable and requires higher participant counts and further analysis. In contrast, our preference was for quality over quantity.

With a higher participant count, we would utilize non-overlapping groups for the two user studies. While we do not expect a significant difference as the samples used in the first study are not directly used in the second study, there could have been some bias due to familiarizing the individuals briefly. We utilize the samples from the first study as source inputs in the second study, which could cause those participating in both studies to choose differently. On the other hand, we have a 15-day break between the two studies, which can avoid the familiarity bias for the second study. ANOVA results assuming two groups of participants as the independent variables reveal no significant differences between the answers of those participating in both studies and those participating only in the second study. The ANOVA statistics comparing the two groups of users based on each output pair are available in our repository.

We should also note that the personality ratings gathered in the first study only reflect the apparent personalities of the people in short video segments and are not expected to depict the real personalities of the individuals. As our aim is on the apparent personality in short first encounters, we do not investigate whether these apparent cues reflect the self-reported personalities or expert assessments; instead, we investigate if motion transfer can alter personality perception for the average user. One of our future interests is to use such a system to express different personalities in video entertainment and education; for example, what happens if we alter the appearance or motion of a tutor teaching a subject?

\section{Conclusion}
We analyzed the influence of different appearance and motion combinations on the perceived personality. To this end, we carried out two user studies. In the first study, we analyzed the apparent personality traits of a set of short video sequences. Then, we selected the samples that best represent each personality factor's high, neutral, and low traits. Using these samples as a motion transfer network's source and driving inputs, we generated stimuli for our second study. The second study compared TPS's output pairs following different combinations to discover their effect on perceived personality.

Our results suggest that using different combinations for TPS's inputs influences extraversion to the most significant extent; utilizing the driving input helps express this factor in general. When the source and driving inputs utilize opposing trait samples in Case 3, we see that the driving input helps convey high extraversion, and the source input helps express high agreeableness and neuroticism. Using TPS has little influence on perceived openness and conscientiousness. Expressing openness in short sequences is challenging, which can explain the lack of influence on openness. Furthermore, our samples in the first study generally scored high in conscientiousness. They had slight variance since they all included presenters, which could have a diminishing effect on comparing the conscientiousness pairs.

Our results can inspire future work aiming for data-driven personality expression; the motion and appearance of videos can be transformed to express the desired traits, enabling control over nonverbal communication. Motion transfer can enhance informative videos to provide a better learning experience; movie sequences can be improved for better character portrayal. Intelligent assistants can employ motion transfer to express different personalities that help communicate with different user types. Future work can automatically generate personality-specific human motions or appearances, which can benefit from an accurate personality prediction system for labeling. Consequently, designers can adjust a few sliders to generate suitable characters or motions to express the desired personality traits. Note that personality perception depends on many aspects of communication; therefore, a complete analysis of the elements involved in personality expression requires more studies. This work will be one more step toward our goal of better-represented digital personalities.

\section*{Acknowledgments}
This research is supported by The Scientific and Technological Research Council of Turkey (T\"{U}B\.{I}TAK) under Grant No. 122E123. 

\bibliographystyle{abbrv}
\bibliography{references}

\end{document}